\documentclass[preprint,12pt,nonatbib,nopreprintline]{elsarticle}
\usepackage[T1]{fontenc}
\usepackage[utf8]{inputenc}
\usepackage[american]{babel}
\usepackage{lmodern}
\usepackage[letterpaper,margin=1in]{geometry}
\usepackage{csquotes}
\usepackage{amsmath,amssymb,amsfonts}
\usepackage{graphicx}
\usepackage{booktabs,tabularx,multirow,array}
\usepackage{makecell}
\usepackage{microtype}
\usepackage{setspace}
\usepackage{placeins}
\usepackage{caption}
\usepackage{adjustbox}
\usepackage{xurl}
\usepackage[hidelinks]{hyperref}
\usepackage[backend=biber,style=apa,sortcites=true]{biblatex}
\renewcommand{\arraystretch}{1.12}
\newcommand{\mainfigwidth}{0.68\textwidth}

\journal{Journal of Pharmacological and Toxicological Methods}
\hypersetup{pdftitle={Predicting Dose-Response Success from Primary Screening Data},pdfauthor={Sean Lim}}

\begin{document}
\begin{frontmatter}
\title{Predicting Dose–Response Success from Primary Screening Data}

\author[a]{Sean Lim\corref{cor1}}
\ead{sean78623@gmail.com}
\cortext[cor1]{Corresponding author.}
\address[a]{Department of BioSciences, Rice University, 6100 Main Street, Houston, TX 77005-1827, USA}

\begin{abstract}
High-throughput drug screening uses inexpensive primary screens to select compounds for more costly follow-up dose--response experiments, which estimate compound potency (pXC$_{50}$) by determining the concentration required to produce a specified biological effect.
However, many promoted compounds fail to yield a reportable potency, even when their primary-screen activity is genuine.
We investigated whether inexpensive three-point primary-screening data could predict whether a more costly 11-point dose--response experiment would yield a reportable pXC$_{50}$.
Using a dataset of 32,971 compound--assay pairs with three-point primary-screening, 11-point dose--response measurements, we trained random-forest models to predict successful potency quantification.
Our model achieved an AUROC of {\unboldmath $0.939 \pm 0.002$} and generalized to previously unseen chemical scaffolds and assay-mechanism families. 
We found that primary-screen response features were more predictive of dose--response quantifiability than molecular structure-based features. 
Successful potency quantification was also strongly associated with primary-screen response amplitude.
Across eight chronological test releases, the model retained 90\% of reportable potencies while selecting only 55.7\% of historically promoted compound--assay pairs for follow-up profiling.
These findings suggest that predicting potency quantifiability from primary-screening data could improve the efficiency of compound prioritization and reduce unnecessary dose--response experiments.
\end{abstract}

\end{frontmatter}

\section{Introduction}\label{sec1}
 
Compound potency is a central measure in early drug discovery. Structure--activity relationships, selectivity profiles, and compound progression commonly depend on potency estimates derived from dose--response experiments, typically reported as IC$_{50}$ or EC$_{50}$ values or as their negative logarithms, such as pXC$_{50}$ \parencite{Hughes2011,SrinivasanLloyd2024}. Obtaining these values, however, requires multi-concentration experiments that are substantially more resource-intensive than the primary screens used to identify candidate hits. 
As a result, high-throughput screening typically follows a multi-stage process in which a large number of compounds are first evaluated using a relatively inexpensive primary assay. 
Only a small subset of promising compound–assay pairs is then selected for more resource-intensive follow-up dose–response experiments \parencite{Hughes2011,Buterez2023,Macarron2011,Inglese2006}. 
A key challenge is therefore deciding which compound–assay pairs should advance to this next stage.

This promotion decision is conventionally made by assessing whether the primary-screen signal represents genuine biological activity, noise, or assay interference. Existing hit-triage approaches accordingly emphasize activity prediction from chemical or historical screening information \parencite{Riniker2014,Buterez2023,Mayr2018}, identification of frequent hitters and assay-interfering compounds \parencite{Baell2010,Dahlin2015,Tan2024}, and replicate-aware statistical methods that distinguish reliable hits from experimental variation \parencite{Malo2010}. More recent machine-learning approaches have further improved true-hit prioritization and false-positive identification \parencite{Boldini2024,Hesse2024}. Although these methods differ substantially, they share a common aim: reducing false-positive activity calls before expensive follow-up. This framing implicitly treats inactivity as the main reason that follow-up profiling fails to yield a useful potency estimate.
 
In the screening campaign analyzed here, this assumption holds only partially. Of 32{,}971 compound--assay pairs promoted to follow-up, 18{,}299 (55.5\%) did not yield a quantifiable potency. Inactivity accounted for just 11{,}549 (63.1\%) of these unsuccessful experiments. The remaining 6{,}750 (36.9\%) were classified as active in the 11-point dose--response experiment, but their concentration--response profiles did not yield a reportable pXC$_{50}$. In other words, even a perfect activity filter would have left more than one-third of the unsuccessful follow-up experiments in place.
 
Thus, the problem is not always that the primary screen identifies a false hit. In many cases, the activity is real, but the response is too weak or incomplete for the follow-up dose--response experiment to produce a reliable Hill fit and a reportable pXC$_{50}$. This explanation is consistent with established principles of dose--response analysis: a curve midpoint can be estimated reliably only when the tested concentrations adequately define the response transition and its asymptotes, and its uncertainty grows when they do not \parencite{Sebaugh2011,Shockley2015}. This observation changes the triage objective. 
Rather than predicting only whether a compound–assay pair will be active, triage should also consider whether follow-up dose–response testing is likely to produce a reliable potency estimate. 
Some compounds may show activity in the primary screen but produce responses that are too weak to support reliable potency estimation in follow-up experiments. In such cases, the primary-screen measurements may already provide evidence that successful potency quantification is unlikely.
In that case, quantifiability, whether a follow-up dose--response experiment yields a reportable pXC$_{50}$, might be predicted before follow-up from data already collected in the standard screening workflow, without additional experiments.
 
Machine learning is increasingly applied to drug-discovery prediction tasks \parencite{Vamathevan2019}, but its usefulness for triage depends on what a model is trained to predict.
Here, we formulate quantifiability as the triage target and test whether it can be predicted from primary-screen data before follow-up. 
We evaluate this formulation on 32{,}971 promoted compound--assay pairs spanning Data Releases 1--11 and five assay-mechanism families of the EvE Bio compound--assay matrix \parencite{EvEBioPortal2026}. 
We report four main findings. 
First, quantifiability is predictable, primarily from the primary-screen response rather than from molecular structure. 
A random forest using 16 simple features of the three-point response outperformed ranking by mean or maximum screening activity, whereas a model using Morgan fingerprints alone performed worse and lost most of its predictive signal under a scaffold-disjoint holdout. 
Second, quantifiability is strongly associated with response amplitude, but this relationship depends on assay context: among pairs active in follow-up, the fraction without a quantifiable potency ranged from 18.8\% to 81.6\% across assay-mechanism families. 
Third, under a release-ordered evaluation in which each release was predicted from earlier releases only, the direct forest retained 90\% of quantifiable potencies while profiling 55.7\% of promoted pairs, compared with 62.2\% for ranking by the reported screening maximum. 
Fourth, the association between screening amplitude and subsequent quantifiability was also observed in an independent, predominantly single-concentration screening dataset, U.S.\ EPA ToxCast/Tox21 \parencite{EPAInvitrodb2025}.

\section{Methods}\label{sec:methods}

\subsection{Study population and outcome definitions}
We performed a retrospective secondary analysis of EvE Bio Data Releases 1--11 \parencite{EvEBioPortal2026}. The analysis used a local snapshot of the release-11 data bundle dated 15 May 2026. Release identifiers in that snapshot were 1.3, 2.2, 3.2, 4.0, 5.1, 6.0, 7.0, 8.0, 9.0, 10.0, and 11.0. The analysis included 32{,}971 compound--assay pairs that had undergone both three-point primary screening and subsequent 11-point dose--response profiling. Each promoted compound--assay pair was included once in the analysis. A compound tested in different assays contributed separate pairs. Because only historically promoted pairs were analyzed, the study estimates performance within that selected population rather than among all screened compounds.

The primary outcome was \emph{quantifiability}: whether the completed profiling experiment yielded a reportable potency estimate under EvE Bio's reporting criteria. For pair $i$,
\begin{equation}
y_i=\begin{cases}
1 & \text{if a reportable pXC$_{50}$ was obtained},\\
0 & \text{otherwise}.
\end{cases}
\end{equation}
Pairs whose source field \texttt{Final\_Category} was \texttt{Active - Quantified (11pt)} were assigned $y_i=1$; all other promoted pairs were assigned $y_i=0$. Three-point features were computed from observed points in the Screening (3pt) phase. The reported screening maximum was the screening-phase value supplied for that pair, not the maximum recorded in the 11-point experiment. For descriptive analyses, the latter group was divided into pairs classified as inactive during profiling and pairs classified as active without a reportable potency. The Active-Poor annotation was used only to characterize the latter group, not as a predictor. We denote activity in the 11-point experiment by $A_i$; under these definitions, $y_i=1$ implies $A_i=1$.

\subsection{Screening-stage predictors and historical context}
Molecular structure, scaffolds, and physicochemical descriptors were calculated with RDKit 2026.03.2 \parencite{RDKit2026,RogersHahn2010}. The response representation comprised 16 features derived from the primary screen: mean, median, maximum, minimum, range, and standard deviation of activity; mean activity at the lowest and highest screening concentrations, $\bar{a}(c_{\min})$ and $\bar{a}(c_{\max})$; their difference, $\bar{a}(c_{\max})-\bar{a}(c_{\min})$; mean and maximum replicate standard deviation; numbers of measurements and unique concentrations; indicators for maximum activity exceeding 30\% or 50\%; and the reported screening maximum. The difference between responses at the highest and lowest concentrations was a response-change proxy, not a fitted Hill slope. Molecular structure was encoded using 1{,}024-bit Morgan fingerprints with radius 2.

Historical context comprised six compound-level screening summaries: activity count, activity fraction, mean and maximum screening amplitude, and numbers of distinct target classes and mechanism families screened. Three assay-level summaries captured activity count, activity fraction, and mean screening amplitude. Mechanism family, target class, mode, technology, screening category, promotion reason, and mutant status were represented by dummy variables. These blocks formed the full 1{,}074-feature representation, $x_i$.

Per-pair predictors used only information recorded before 11-point profiling (Supplementary Table S1). The 16 three-point features, the reported screening maximum, Morgan fingerprints, assay descriptors, screening category, and promotion reason are screening-stage fields. Eleven-point measurements defined the outcomes and described profiling responses. They were not model inputs. In the random-split analyses the nine historical summaries were calculated from the full screening matrix. In the release-ordered evaluation they were recomputed from earlier releases.

\subsection{Prediction models and screening comparators}
The principal direct predictor was a class-balanced random forest with 500 trees, trained to predict $y_i$ from the full feature representation \parencite{Breiman2001}. Class-balanced logistic regression using standardized full-feature values provided a linear comparator. Additional 300-tree class-balanced random forests used either the 16 response features or the molecular fingerprints alone; each was fitted on the full training portion of its corresponding split.

The non-ML comparators ranked pairs by mean primary-screen activity, maximum primary-screen activity, or random order. Mean activity was the principal heuristic in the five-fold random comparison. Maximum activity was the comparator for the chronological profiling-budget analyses. These two screening comparators are therefore distinguished throughout the evaluation.

Screening category and promotion reason are screening-stage annotations provided by EvE Bio. The full 500-tree forest was therefore refit without their seven and three indicator columns (1{,}064 features) on the same folds.

Isolated feature blocks and nested feature addition used the same five-fold stratified cross-validation and 300-tree forests. Nested models successively added fingerprints, historical profiles, and assay metadata to the three-point block. Grouped feature importance was the sum of impurity-based importance within each block.

\subsection{Software, seeds, and fitting settings}
Models were fit with scikit-learn 1.8.0 \parencite{Pedregosa2011}. Data processing used NumPy 2.3.5 and pandas 2.3.3, and hypothesis testing used SciPy 1.17.1. Every fitted model and every shuffled partition used random seed 42. No hyperparameter search was performed.

Random forests used unrestricted depth and balanced class weights. The direct full-feature model and the forest used for grouped importance had 500 trees. Feature ablations, scaffold-disjoint and mechanism-held-out forests, each stage of the two-stage score, the matrix-completion comparator, and the plate-control comparison had 300 trees. Logistic regression used L2 regularization with inverse strength $C=1$, the lbfgs solver, at most 1{,}000 iterations, and balanced class weights. Its predictors were standardized with a scaler fit inside the training fold. The subgroup gradient-boosted tree was scikit-learn's gradient-boosting classifier with 500 trees, maximum depth 5, learning rate 0.05, row subsampling fraction 0.8, and the library's default binary classification loss. The area under the precision--recall curve was average precision.

\subsection{Random, scaffold-disjoint, and mechanism-held-out evaluation}
The primary comparison used five-fold stratified random cross-validation at the compound--assay-pair level. The folds did not hold out compound identities or assays. To assess transfer across chemical series, we used five-fold Bemis--Murcko scaffold-disjoint splitting across 787 scaffolds, with 300-tree forests \parencite{BemisMurcko1996,WallachHeifets2018}. Compounds assigned to the same scaffold were kept in the same fold. A SMILES string that did not yield a scaffold received its own compound-specific placeholder, so those compounds were not grouped together.

Transfer across assay-mechanism families was assessed by holding out each of the five families in turn. The compared representations were fingerprints alone, fingerprints plus the three-point block, and fingerprints plus the three-point block and mechanism-family indicators. This evaluation tests transfer to an omitted mechanism family within the same data source; it is not an independent-laboratory validation.

\subsection{Release-ordered evaluation}
We evaluated later releases with a model trained only on earlier releases \parencite{Sheridan2013}. Releases 1.3, 2.2, and 3.2 formed the initial training history. Each of releases 4.0, 5.1, 6.0, 7.0, 8.0, 9.0, 10.0, and 11.0 was then evaluated using the direct 500-tree forest trained on promoted pairs from earlier releases, giving eight test windows containing 30{,}120 pairs. Release 5.1, rather than an unobserved 5.0, was the next included release after release 4.0. For each window, the nine historical features were recomputed from screening results preceding that test release; unseen compounds or assays were assigned training-history priors.

Chronological AUROC, recall at a 50\% budget, and $B(0.90)$ were computed from that same forest. The baseline was ranking by the reported screening maximum. Rankings were calculated separately within each test release. A two-stage score, $\hat{P}(A_i=1\mid x_i)\hat{P}(y_i=1\mid A_i=1,x_i)$ with 300 trees at each stage, changed five-fold AUROC by 0.001 relative to the direct forest and was not used for these results.

\subsection{Discrimination, calibration, and profiling-budget metrics}
Discrimination was summarized using the area under the receiver operating characteristic curve (AUROC) and the area under the precision--recall curve (AUPRC), the latter computed as average precision. Where cross-validation summaries are reported, values are means and standard deviations across folds. For the full-model AUROC and its paired differences from mean three-point activity and from the annotation-free forest, 95\% percentile confidence intervals were obtained from 2{,}000 compound-cluster bootstrap resamples of the out-of-fold predictions, resampling the 1{,}075 compounds with replacement.

For a test set of $n$ pairs, let $\pi$ order pairs from highest to lowest score and let $k=\lfloor fn\rfloor$ for a profiling fraction $f$. With at least one quantifiable pair and $k\geq1$, recall and yield are
\begin{equation}
\mathrm{Recall}@f=\frac{\sum_{i=1}^{k}y_{\pi(i)}}{\sum_{j=1}^{n}y_j},
\qquad
\mathrm{Yield}@f=\frac{1}{k}\sum_{i=1}^{k}y_{\pi(i)}.
\end{equation}
Recall measures the fraction of all reportable-potency outcomes recovered, whereas yield measures the fraction of selected pairs yielding reportable potency. We report Recall@50\% and Yield@25\%, consistent with evaluating early retrieval in ranked screening \parencite{TruchonBayly2007}.

The profiling fraction required to recover a proportion $r$ of quantifiable outcomes in a completed test set is
\begin{equation}
B(r)=\min_{k\in\{1,\ldots,n\}}
\left\{\frac{k}{n}:\sum_{i=1}^{k}y_{\pi(i)}\geq
r\sum_{j=1}^{n}y_j\right\},
\end{equation}
with $r=0.90$ in the chronological analysis. The aggregate profiling budget was calculated by summing the numbers of top-ranked pairs required to reach 90\% recall within each release and dividing this total by the 30{,}120 pairs across the eight test windows.

Expected calibration error and the Brier score,
\begin{equation}
\mathrm{BS}=\frac{1}{n}\sum_{i=1}^{n}(\hat{s}_i-y_i)^2,
\end{equation}
were calculated from the direct forest's out-of-fold scores $\hat{s}_i$ on the promoted set. Expected calibration error used ten equal-width bins on $[0,1]$, each weighted by its share of pairs.

\subsection{Exploratory analyses of signal and errors}
Physicochemical descriptors---molecular weight, LogP, hydrogen-bond donor and acceptor counts, rotatable-bond count, and topological polar surface area---were compared between quantifiable and active-unquantifiable groups with two-sided Welch tests, across pairs and across unique compounds with at least one pair in a group. Effect sizes are Cohen's $d$ with the pooled standard deviation. Because multiple compound--assay pairs could involve the same compound, pair-level confidence intervals for $d$ were estimated using 2{,}000 compound-cluster bootstrap resamples. No multiplicity adjustment was applied. Assay-level control summaries were derived from 516{,}894 control wells across 16{,}212 plates \parencite{Zhang1999}. Within each plate, activity was summarized separately for the minimum-control and maximum-control wells. The signal window was the difference between those two means. A $Z'$-factor proxy was
\begin{equation}
Z' = 1 - \frac{3(\mathrm{SD}_{\max}+\mathrm{SD}_{\min})}{\max(w,\,0.01)},
\end{equation}
where $w$ is the signal window. Plate values were then averaged within assay. The five resulting predictors were the mean and standard deviation of $Z'$, the mean signal window, and the mean standard deviations of the minimum and maximum controls. Missing assay-level values were filled with zero, and the five columns were appended to the full representation for the 300-tree comparison.

The matrix-completion comparator was a 300-tree class-balanced random forest under the same five-fold stratified splits. Its predictors were the 1{,}024 fingerprint bits and the assay descriptors for target class, mode, mechanism family, technology, and mutant status. The focal pair's 16 three-point measurements were omitted, as were the historical activity profiles. The interpretable comparator also omitted fingerprints. It used the reported screening maximum, indicators for mechanism family and target class, and the product of the screening maximum with each of those indicators. Feature contributions were summarized using impurity-based importance scores. The interaction share is the summed importance of the product columns.

Subgroup discrimination and within-subgroup recall were evaluated with the gradient-boosted tree specified above, trained on the complement of one stratified fold and scored on the held-out fold of 6{,}595 pairs. The missed-pair analysis used a 500-tree random forest and a 50\% profiling budget over that entire held-out fold, rather than separate budgets within each subgroup.

\subsection{Independent assessment in ToxCast/Tox21}
We examined U.S.\ EPA ToxCast/Tox21 invitrodb version 4.3, the August 2025 release \parencite{EPAInvitrodb2025,Feshuk2023}. Multi-concentration series were the representative sample rows (\texttt{chid\_rep} $= 1$) of the level-5 curve-model table, joined to level-4 summary rows on the level-4 identifier and the assay-component-endpoint identifier. A series was labeled quantifiable when the absolute hit call was at least 0.9, the selected curve model was not constant, and at least one potency parameter among AC50, ACC, BMD, and the gain term was finite. Screening rows were the representative level-2 screening summaries, joined to those series on assay-component endpoint and sample identifier. The ranking score was the maximum absolute median response across screening concentrations, so both directions of activity were treated as amplitude. When that concentration-level summary was missing, the absolute level-2 maximum median was used. Pairs missing the score or the quantifiable label were excluded.

Merging the tables yielded 43{,}242 compound--endpoint pairs across 361 endpoints. Of these, 42{,}479 (98.2\%) had a single screening concentration, so the score was the absolute screening amplitude and the EvE Bio model was not applied. The amplitude score was evaluated under five-fold stratified cross-validation, shuffled with random seed 0, and under five-fold group cross-validation that held out assay-component endpoints.

\section{Results}\label{sec2}

\begin{figure}[!htbp]
\centering
\includegraphics[width=\textwidth,keepaspectratio]{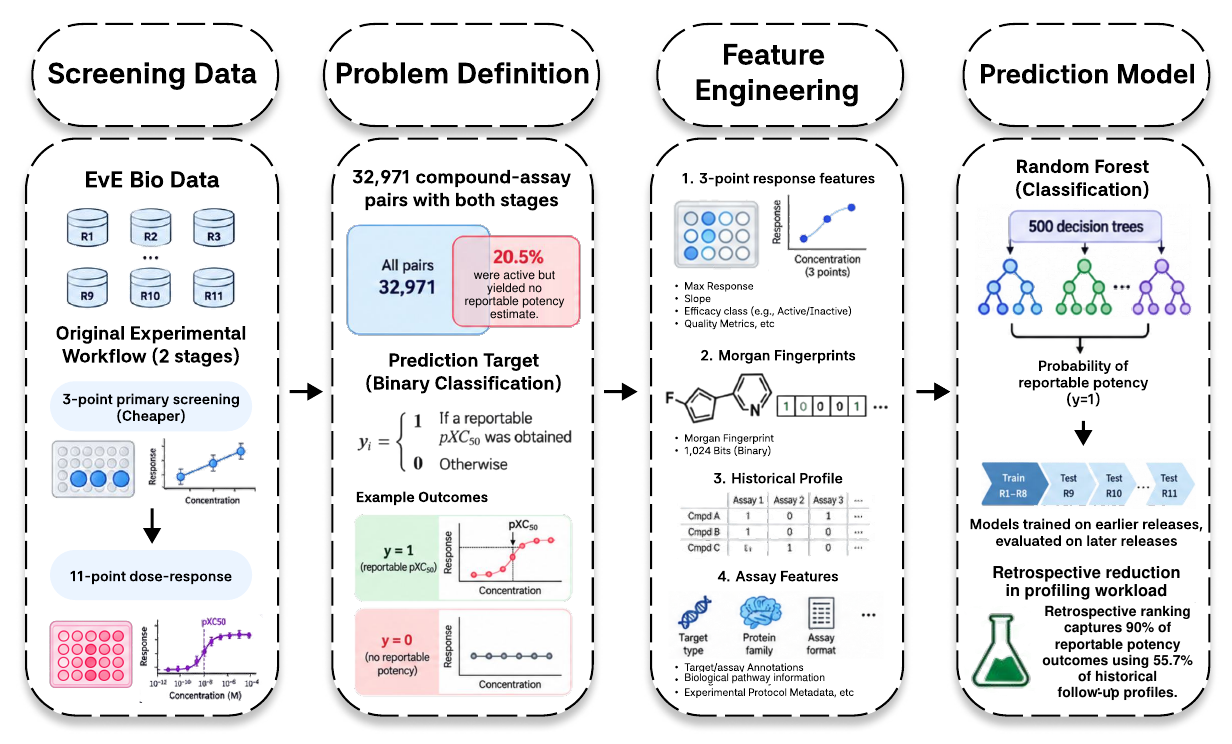}
\caption{Study design and retrospective evaluation workflow. Historically promoted compound--assay pairs with three-point primary-screen data and follow-up 11-point dose--response profiling were used to train models that predict whether follow-up profiling yields a reportable pXC$_{50}$. The retrospective release-ordered evaluation compares quantifiability-aware ranking with a primary-screen-activity baseline and summarizes the profiling budget required to recover reportable potency outcomes.}
\label{fig:study_design}
\end{figure}

Figure~\ref{fig:study_design} summarizes the overall study design and evaluation workflow. We analyzed 32{,}971 historically promoted compound--assay pairs from EvE Bio, using features available after the three-point primary screen to predict whether the subsequent 11-point dose--response experiment would yield a reportable pXC$_{50}$. We then asked whether ranking pairs by predicted quantifiability could improve retrospective prioritization of follow-up profiling relative to ranking by primary-screen activity alone.

\subsection{Quantifiability can be predicted from the three-point primary screen}

\begin{figure}[!htbp]
\centering
\includegraphics[width=\mainfigwidth]{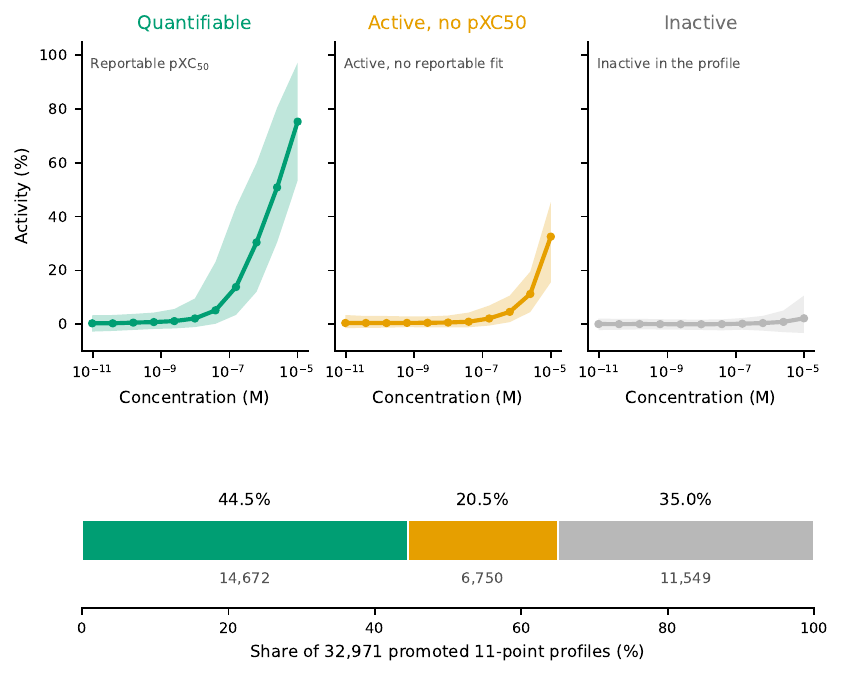}
\caption{Outcomes of the 32{,}971 promoted 11-point profiles. Top: median activity (line) and interquartile range (band) at each of the 11 concentration positions, plotted at the median concentration of that position. Bottom: share and number of promoted pairs in each outcome.}
\label{fig:outcomes}
\end{figure}

Of the 32{,}971 pairs promoted to an 11-point experiment, 14{,}672 (44.5\%) yielded a reportable potency, 6{,}750 (20.5\%) were active without one, and 11{,}549 (35.0\%) were inactive (Fig.~\ref{fig:outcomes}). Among pairs active during profiling, 31.5\% did not yield a potency estimate. We therefore asked whether quantifiability could be predicted from information already available after the three-point screen.

\begin{table}[!htbp]
\centering
\small
\caption{Five-fold random cross-validation performance for predicting whether a promoted compound--assay pair yields a reportable potency. Values are mean $\pm$ standard deviation. Recall@50\% is the fraction of quantifiable pairs recovered when the top-ranked 50\% of pairs are profiled. Yield@25\% is the fraction of profiled pairs that are quantifiable when the top-ranked 25\% are selected. Fingerprint-only and three-point-only random forests are class-balanced 300-tree models fit on the full training fold of each split, matching the isolated-block protocol in Table~\ref{tab:feature_ablation}. Logistic regression and the full random forest use the complete screening-stage feature set; the full forest uses 500 trees. The annotation-free forest is the same 500-tree forest without the screening-category and promotion-reason indicators.}
\label{tab:cv_performance}
\setlength{\tabcolsep}{2.5pt}
\begin{tabular}{lcccc}
\toprule
Policy & AUROC & AUPRC & Recall@50\% & Yield@25\% \\
\midrule
Random & $0.502 \pm 0.003$ & $0.449 \pm 0.005$ & $0.500 \pm 0.003$ & $0.450 \pm 0.009$ \\
Random forest (fingerprints only) & $0.761 \pm 0.004$ & $0.690 \pm 0.007$ & $0.711 \pm 0.003$ & $0.720 \pm 0.006$ \\
Maximum three-point activity & $0.848 \pm 0.003$ & $0.781 \pm 0.005$ & $0.807 \pm 0.005$ & $0.850 \pm 0.006$ \\
Mean three-point activity & $0.863 \pm 0.002$ & $0.829 \pm 0.003$ & $0.814 \pm 0.002$ & $0.888 \pm 0.003$ \\
Random forest (three-point only) & $0.886 \pm 0.002$ & $0.867 \pm 0.002$ & $0.834 \pm 0.005$ & $0.908 \pm 0.003$ \\
Logistic regression (full features) & $0.913 \pm 0.004$ & $0.889 \pm 0.005$ & $0.869 \pm 0.005$ & $0.932 \pm 0.007$ \\
\textbf{Random forest (full features)} & $\mathbf{0.939 \pm 0.002}$ & $\mathbf{0.927 \pm 0.002}$ & $\mathbf{0.894 \pm 0.004}$ & $\mathbf{0.963 \pm 0.003}$ \\
Random forest (no annotations) & $0.939 \pm 0.002$ & $0.927 \pm 0.002$ & $0.895 \pm 0.003$ & $0.960 \pm 0.004$ \\
\bottomrule
\end{tabular}
\end{table}

Under five-fold random cross-validation, simple screening activity was already informative: ranking pairs by maximum three-point activity achieved an AUROC of $0.848 \pm 0.003$, while mean three-point activity achieved $0.863 \pm 0.002$. We used the latter as the principal screening baseline.

We extracted 16 features from the three-point screen, all available before 11-point profiling: amplitude statistics (mean, median, maximum, minimum, range, and standard deviation), activity at the lowest and highest screening concentrations, a slope proxy, replicate-noise summaries, the number of measurements and unique concentrations, binary flags for strong activity, and the reported screening maximum. We then combined this block with 1{,}024-bit Morgan fingerprints of the compound, historical compound- and assay-level screening profiles (how often the compound had been active across assays, and how often the assay had yielded activity across compounds), and assay metadata encoding mechanism family, target class, technology, screening category, promotion reason, and a mutant flag.

Logistic regression on the full screening-stage set reached $0.913 \pm 0.004$. A random forest trained on the full set achieved an AUROC of $0.939 \pm 0.002$ (95\% CI $0.932$--$0.946$) and an AUPRC of $0.927 \pm 0.002$, recovering 89.4\% of quantifiable pairs at a 50\% profiling budget (Table~\ref{tab:cv_performance}). The full model's AUROC exceeded that of ranking by mean three-point activity by $0.076$ (95\% CI $0.069$--$0.083$). Removing screening category and promotion reason, the two provider-assigned annotations among the predictors, left the AUROC at $0.939 \pm 0.002$; the paired difference was $0.0005$ (95\% CI $-0.0001$ to $0.0011$).

Performance varied less across model classes than across feature sets, suggesting that predictive performance was driven primarily by information contained in the screening data rather than by the choice of learning algorithm.

\subsection{Predictive information is concentrated in the screening response rather than molecular structure}

We next examined which sources of information accounted for the predictability of quantification. Holding the model class fixed, a random forest using just the 16 features derived from the three-point screen achieved an AUROC of $0.886 \pm 0.002$, whereas the same protocol using 1{,}024-bit Morgan fingerprints alone achieved $0.761 \pm 0.004$ (Tables~\ref{tab:cv_performance} and~\ref{tab:feature_ablation}). The three-point block therefore exceeded ranking by mean three-point activity ($0.863 \pm 0.002$) without using molecular structural information.

\begin{table}[!htbp]
\centering
\small
\caption{Feature-group ablation for a random forest trained on the quantified label
($n_{\mathrm{estimators}}=300$, class-balanced). Isolated blocks were evaluated by
five-fold stratified cross-validation. Nested models added groups cumulatively and
were evaluated by the same five-fold stratified cross-validation. The three-point block
contains 15 response summaries plus the reported screening maximum (16 features).
Compound and assay profiles comprise six compound-level and three assay-level
historical screening statistics (9 features). The final nested step restores assay
metadata, mechanism family, target class, technology, screening category, promotion
reason, and a mutant flag, to recover the full 1{,}074-feature matrix. Share of
nested gain is the fraction of the AUROC increase from the three-point block
($0.886$) to the full model ($0.939$). Isolated five-fold values use the same
300-tree, full-training-fold protocol as the corresponding single-block rows in
Table~\ref{tab:cv_performance}.}
\label{tab:feature_ablation}

\begingroup
\footnotesize
\setlength{\tabcolsep}{1.8pt}
\renewcommand{\arraystretch}{1.08}

\begin{tabular}{@{}llrcccc@{}}
\toprule
Feature set & Contents & $N$ & AUROC & AUPRC & Rec@50\% & $\Delta$AUROC \\
\midrule

\multicolumn{7}{@{}l}{\textit{Isolated blocks (five-fold)}} \\

Three-point only
& 3pt block
& 16
& $0.886 \pm 0.002$
& $0.867 \pm 0.002$
& $0.834 \pm 0.005$
& --- \\

Fingerprints only
& Morgan bits
& 1{,}024
& $0.761 \pm 0.004$
& $0.690 \pm 0.007$
& $0.711 \pm 0.003$
& --- \\

Full model
& all blocks
& 1{,}074
& $0.939 \pm 0.002$
& $0.926 \pm 0.002$
& $0.894 \pm 0.004$
& --- \\

\midrule

\multicolumn{7}{@{}l}{\textit{Nested addition (five-fold)}} \\

Three-point only
& 3pt block
& 16
& $0.886 \pm 0.002$
& $0.867 \pm 0.002$
& $0.834 \pm 0.005$
& --- \\

+ fingerprints
& + Morgan bits
& 1{,}040
& $0.919 \pm 0.003$
& $0.904 \pm 0.003$
& $0.871 \pm 0.004$
& $+0.033$ \\

+ profiles
& + compound/assay history
& 1{,}049
& $0.930 \pm 0.002$
& $0.916 \pm 0.003$
& $0.885 \pm 0.004$
& $+0.011$ \\

+ metadata (full)
& + assay metadata
& 1{,}074
& $0.939 \pm 0.002$
& $0.926 \pm 0.002$
& $0.894 \pm 0.004$
& $+0.009$ \\

\bottomrule
\end{tabular}

\endgroup
\end{table}

\begin{figure}[!htbp]
\centering
\includegraphics[width=\mainfigwidth]{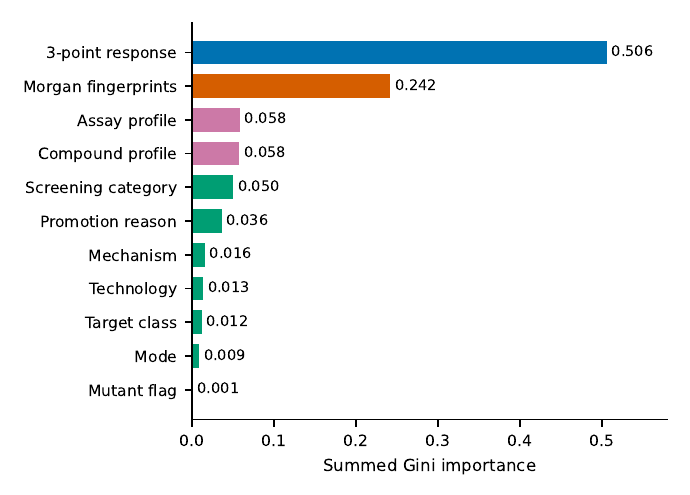}\\[0.4em]
\includegraphics[width=\mainfigwidth]{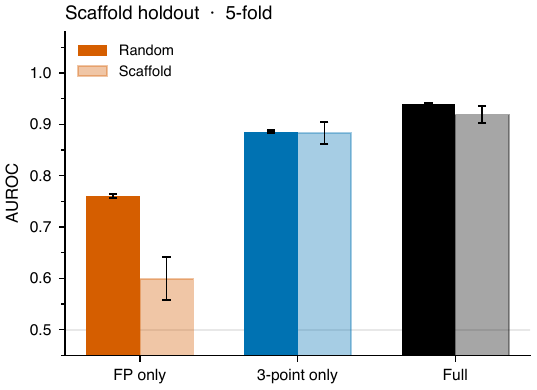}
\caption{Predictive information is concentrated in the screening response. Top: summed Gini importance by feature group for a random forest trained on the full screening-stage matrix. The 16 three-point features account for the largest share, followed by Morgan fingerprints; historical profiles and assay metadata contribute smaller amounts. Bottom: AUROC under random versus Bemis--Murcko scaffold-disjoint splitting for fingerprint-only, three-point-only, and full models. Bars are five-fold means; error bars are one standard deviation. The three-point block is nearly unchanged on unseen scaffolds, whereas fingerprint-only performance declines substantially.}
\label{fig:features}
\end{figure}

Nested feature-group ablation on the same five folds showed the same ordering (Table~\ref{tab:feature_ablation}). 
Starting from the three-point block, adding Morgan fingerprints produced the largest subsequent gain ($\Delta$AUROC $+0.033$; 63\% of that gain). 
Historical compound and assay profiles added $+0.011$, and residual assay metadata added $+0.009$. 
Isolated fingerprints under the same forest reached an AUROC of $0.761 \pm 0.004$, well below the three-point block alone. 
Thus molecular structure and historical context contributed additional information, but the largest predictive component was already contained in the three-point response (Fig.~\ref{fig:features}).

The gap widened under scaffold-disjoint evaluation, in which test compounds share no Bemis--Murcko scaffold with training compounds. 
A Bemis--Murcko scaffold is the core ring system of a molecule, so this split places entire chemical series in either the training or the test set and asks whether the model can predict for specific chemotypes it has not seen.
The three-point-only model's AUROC was nearly unchanged, decreasing from $0.886 \pm 0.002$ to $0.883 \pm 0.021$. 
In contrast, the fingerprint-only model's AUROC decreased from $0.761 \pm 0.004$ to $0.599 \pm 0.042$ (Fig.~\ref{fig:features}). 
Screening-response features therefore retained most of their predictive performance on unseen scaffolds, whereas structural features were substantially less stable under chemotype separation.

We also tested whether quantifiability could be predicted without the three-point response of the pair being ranked, using only the compound's fingerprint and the assay's descriptors (target class, mode, mechanism family, technology, and mutant status).
A matrix-completion model using fingerprints and assay descriptors but no three-point measurements reached an AUROC of 0.842, compared with 0.939 for the full per-pair model. 
Together, these analyses identify the observed primary screening response as the dominant transferable source of information for predicting whether a later dose--response profile will yield a reportable potency.

\subsection{Active but unquantifiable profiles are characterized by weak response amplitude}

We next examined what distinguishes active responses that yield a reportable potency from active responses that do not. 
Bulk physicochemical properties provided little separation between the two groups (Supplementary Table S2). 
At the pair level, compounds in the active-unquantifiable group tended to be slightly larger, more polar, and more flexible. 
For molecular weight, hydrogen-bond donors, rotatable bonds, and topological polar surface area, the confidence interval for Cohen’s d did not include zero, indicating a small but detectable difference between groups. 
Every effect was small ($|d| < 0.17$; the largest interval bound was $0.25$, for rotatable bonds). 
Across the 726 and 671 compounds represented in the two groups, none of the six physiochemical properties differeed significantly between groups ($p \geq 0.17$; $|d| \leq 0.07$). 
Adding five assay-quality features derived from 516{,}894 control wells across 16{,}212 plates raised AUROC only slightly, from $0.939 \pm 0.002$ to $0.943 \pm 0.002$.

\begin{figure}[!htbp]
\centering
\includegraphics[width=\mainfigwidth]{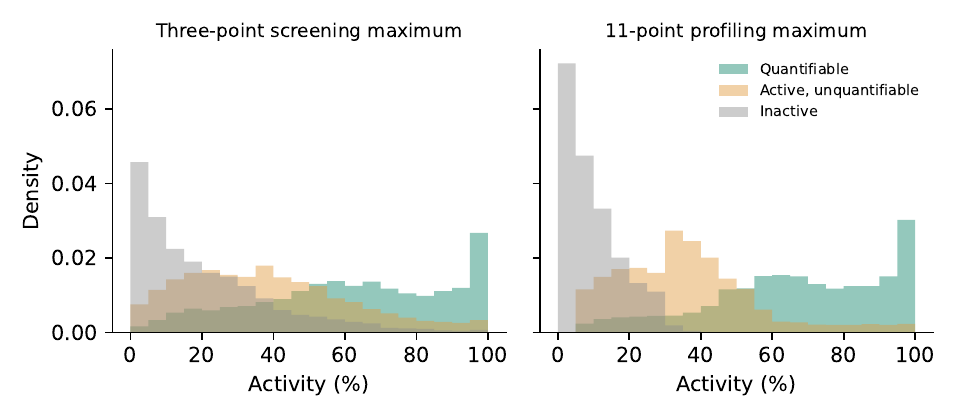}
\caption{Distributions of observed maximum activity by profiling outcome for the 32{,}971 promoted pairs. Left: reported screening maximum from the three-point screen. Right: maximum activity in the 11-point profile. Quantifiable pairs ($n = 14{,}672$), active but unquantifiable pairs ($n = 6{,}750$), and inactive pairs ($n = 11{,}549$). Means are 68.0\%, 42.6\%, and 19.0\% in the screen and 75.9\%, 52.0\%, and 8.6\% in the profile.}
\label{fig:amplitude}
\end{figure}

Response amplitude, in contrast, differed strongly across the three profiling outcomes (Fig.~\ref{fig:amplitude}). 
Mean maximum activity observed in the three-point primary screen was 68.0\% for pairs that later yielded a quantifiable potency, 42.6\% for active but unquantifiable pairs, and 19.0\% for pairs that were inactive during profiling. 
The separation between quantifiable and active-unquantifiable pairs remained visible in the subsequent 11-point experiment, where maximum activity averaged 75.9\% and 52.0\%, respectively. 
Weak responses were particularly enriched among active profiles that failed quantification. 
Among active-unquantifiable pairs, 59.9\% had a maximum profiling response below 40\%, compared with 11.9\% of quantifiable pairs.

The relationship between amplitude and quantification differed substantially across assay-mechanism families. Among pairs confirmed active in the 11-point experiment, Hill-fitting failure ranged from 18.8\% in G-protein activation assays to 81.6\% in heterodimer assays (Table~\ref{tab:mechanism_fail}).

\begin{table}[!htbp]
\centering
\small
\caption{Hill-fitting failure among compound--assay pairs that were active in the 11-point experiment, by assay-mechanism family. Failure is the fraction of active pairs that did not yield a reportable potency. Totals match the 21{,}422 pairs confirmed active during profiling, of which 6{,}750 (31.5\%) were unquantifiable.}
\label{tab:mechanism_fail}
\begin{tabular}{lrrrr}
\toprule
Mechanism family & $N$ active & Quantifiable & Unquantifiable & Failure (\%) \\
\midrule
G-protein activation & 6{,}824 & 5{,}542 & 1{,}282 & 18.8 \\
Competition binding & 4{,}199 & 3{,}038 & 1{,}161 & 27.6 \\
Barr2 recruitment & 8{,}375 & 5{,}394 & 2{,}981 & 35.6 \\
Co-factor recruitment & 890 & 489 & 401 & 45.1 \\
Heterodimer & 1{,}134 & 209 & 925 & 81.6 \\
\midrule
All active pairs & 21{,}422 & 14{,}672 & 6{,}750 & 31.5 \\
\bottomrule
\end{tabular}
\end{table}

An interpretable model containing screening amplitude together with assay-mechanism and target-class interactions achieved an AUROC of 0.816 without using molecular fingerprints. 
Interaction terms accounted for 36.8\% of feature importance, with the interaction between assay mechanism and screening maximum providing the largest individual contribution.

Predictive performance also varied across assay subgroups (Table~\ref{tab:subgroup_errors}). 
AUROC reached 0.975 for kinase competition-binding assays, 0.920 for 7TM assays, and 0.944 for nuclear-receptor assays, while screening categories characterized by inconsistent responses were more difficult to rank. 
Residual errors were concentrated among weak responses: at a 50\% profiling budget,  the quantifiable pairs the model missed had much weaker screening responses of 31.1\%, than those it recovered (71.2\%)
Of 316 missed quantifiable pairs, 173 occurred in Barr2 recruitment assays.

\begin{table}[!htbp]
\centering
\small
\caption{Subgroup ranking and residual errors on a held-out fold of 6{,}595 promoted pairs. Target-class and screening-category metrics are from a gradient-boosted tree trained on the remaining pairs. $N$ is the number of held-out pairs in the subgroup; Pos.\ is the fraction that were quantifiable; Rec@50\% is recall when the top-ranked half of that subgroup is profiled. Missed-pair counts and screening maxima are from a random forest evaluated at a 50\% budget over the full held-out fold (2{,}619 of 2{,}935 quantifiable pairs recovered). Kinase and competition-binding rows coincide because the kinase pairs in this dataset are the competition-binding assays.}
\label{tab:subgroup_errors}
\setlength{\tabcolsep}{4pt}
\begin{tabular}{lrrrr}
\toprule
Subgroup & $N$ & Pos.\ (\%) & AUROC & Rec@50\% \\
\midrule
\multicolumn{5}{l}{\textit{Target class}} \\
Kinase (competition binding) & 1{,}547 & 37.8 & 0.975 & 0.974 \\
Nuclear receptor & 704 & 19.9 & 0.944 & 0.964 \\
7TM & 4{,}344 & 50.9 & 0.920 & 0.828 \\
\midrule
\multicolumn{5}{l}{\textit{Screening category}} \\
Compound of interest & 2{,}710 & 73.5 & 0.923 & 0.665 \\
Inconsistent, high & 689 & 48.0 & 0.916 & 0.840 \\
Inactive & 1{,}780 & 7.2 & 0.906 & 0.969 \\
Unmeasurable pXC$_{50}$ & 994 & 35.9 & 0.871 & 0.866 \\
Low activity & 103 & 19.4 & 0.831 & 0.900 \\
Inconsistent, low & 191 & 41.9 & 0.765 & 0.738 \\
Inconsistent, random & 128 & 20.3 & 0.736 & 0.692 \\
\midrule
\multicolumn{5}{l}{\textit{Missed quantifiable pairs at a 50\% budget}} \\
 & $N$ missed & Share of misses (\%) & Mean 3-pt max (\%) & \\
\cmidrule{1-4}
Recovered quantifiable & 2{,}619 & --- & 71.2 & \\
Missed quantifiable & 316 & 100 & 31.1 & \\
\quad Barr2 recruitment & 173 & 54.7 & --- & \\
\quad G-protein activation & 53 & 16.8 & --- & \\
\quad Competition binding & 53 & 16.8 & --- & \\
\quad Co-factor recruitment & 20 & 6.3 & --- & \\
\quad Heterodimer & 17 & 5.4 & --- & \\
\bottomrule
\end{tabular}
\end{table}

Together, these results show that failure to obtain a potency from an active pair is strongly associated with weak or partial response amplitude and that the amplitude associated with successful quantification depends on assay context.

\subsection{Chronological evaluation quantifies the profiling-budget trade-off}

We evaluated the direct forest chronologically across successive EvE Bio releases. 
For each test release from release 4 through release 11, the forest was trained on promoted pairs from earlier releases, and historical compound- and assay-level features were recomputed from that prior history. 
Releases 1--3 were the initial training history, leaving 30{,}120 pairs across eight test windows.

\begin{figure}[!htbp]
\centering
\includegraphics[width=\mainfigwidth]{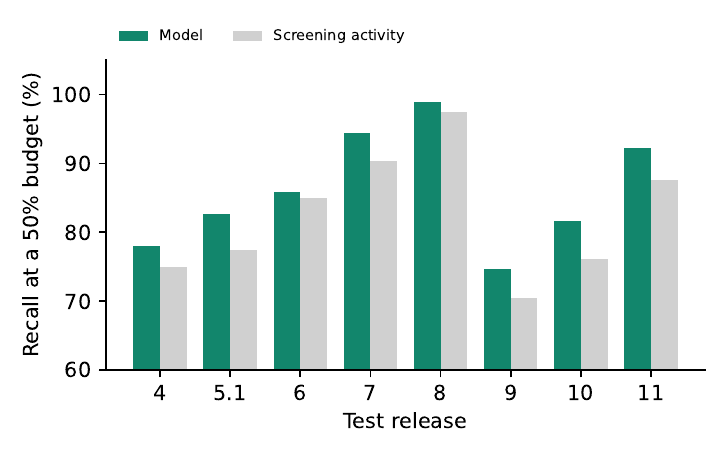}\\[0.4em]
\includegraphics[width=\mainfigwidth]{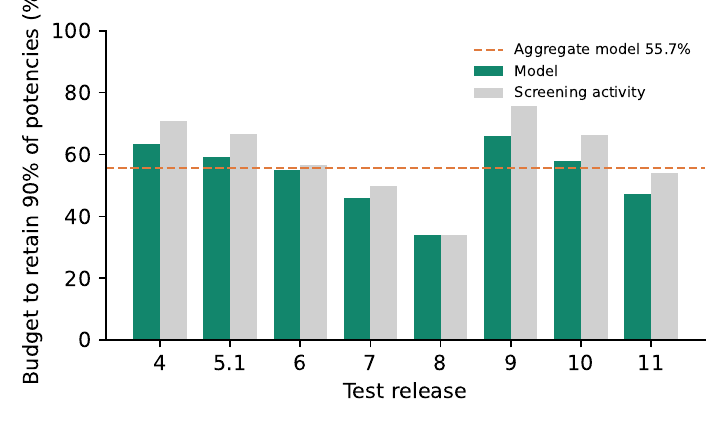}
\caption{Release-ordered evaluation across eight incoming releases, using the direct forest and ranking by the reported screening maximum. Top: fraction of quantifiable pairs recovered when the top-ranked 50\% of each release is profiled. Bottom: fraction of each release profiled to retain 90\% of the potencies obtained by exhaustive profiling. The dashed line is the aggregate model budget of 55.7\%. The model required the smaller budget on every release.}
\label{fig:rolling}
\end{figure}

Under this evaluation, direct-forest AUROC ranged from 0.814 to 0.977, with a median of 0.900. 
At a 50\% profiling budget, the same forest recovered a greater fraction of quantifiable pairs than simply ranking by the reported screening maximum on all eight releases (Table~\ref{tab:temporal}; Fig.~\ref{fig:rolling}).

\begin{table}[!htbp]
\centering
\small
\caption{Release-ordered evaluation with one model and one baseline. For each test release, the direct 500-tree forest was trained only on promoted pairs from earlier releases, and historical features were recomputed from that prior history. AUROC and Rec@50\% were calculated using predictions from this forest. The screening column ranks by the reported screening maximum at the same 50\% budget. Release 5.1 is the first test window after release 4. The eight windows contain 30{,}120 pairs.}
\label{tab:temporal}
\setlength{\tabcolsep}{4pt}
\begin{tabular}{crrcccc}
\toprule
Release & Test $N$ & Quantifiable & Pos.\ (\%) & AUROC & Rec@50\% (model) & Rec@50\% (screening) \\
\midrule
4 & 4{,}731 & 1{,}966 & 41.6 & 0.814 & 0.780 & 0.750 \\
5.1 & 4{,}462 & 1{,}851 & 41.5 & 0.866 & 0.827 & 0.774 \\
6 & 2{,}739 & 1{,}297 & 47.4 & 0.917 & 0.858 & 0.850 \\
7 & 2{,}722 & 1{,}079 & 39.6 & 0.945 & 0.943 & 0.903 \\
8 & 2{,}475 & 805 & 32.5 & 0.977 & 0.989 & 0.974 \\
9 & 3{,}571 & 2{,}077 & 58.2 & 0.883 & 0.747 & 0.704 \\
10 & 7{,}077 & 3{,}942 & 55.7 & 0.940 & 0.817 & 0.761 \\
11 & 2{,}343 & 258 & 11.0 & 0.828 & 0.922 & 0.876 \\
\bottomrule
\end{tabular}
\end{table}

Recall at a fixed profiling fraction depended strongly on the prevalence of quantifiable pairs. 
Release 9, in which 58.2\% of promoted pairs were quantifiable, had an AUROC of 0.883 but recovered 74.7\% of quantifiable pairs at a 50\% budget. 
Release 11 contained only 11.0\% quantifiable pairs and had an AUROC of 0.828, but recovered 92.2\% of quantifiable pairs at the same budget. 
Fixed-budget recall is therefore compared with the screening ranking inside each release.

Across the eight test releases, the direct forest recovered 90\% of the potencies obtained by exhaustive profiling while profiling 55.7\% of promoted pairs (16{,}780 of 30{,}120; Fig.~\ref{fig:rolling}). Ranking by the reported screening maximum required 62.2\% (18{,}724 experiments) for the same retention. The model used 1{,}944 fewer experiments than that baseline, a 10.4\% reduction, and 13{,}340 fewer than profiling every promoted pair. The model budget was smaller on all eight releases, including release 8 (33.9\% versus 34.1\%).

\begin{figure}[!htbp]
\centering
\includegraphics[width=\mainfigwidth]{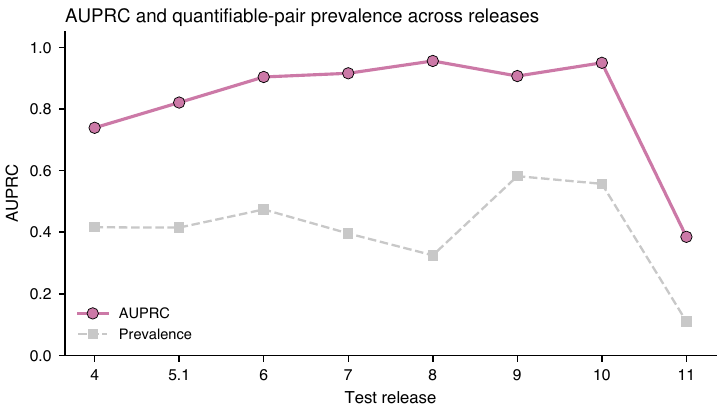}
\caption{Precision--recall performance tracks the prevalence of quantifiable pairs across chronological test releases. AUPRC remains high while the positive rate is moderate, then falls to 0.382 on release 11, where only 11.0\% of promoted pairs were quantifiable.}
\label{fig:auprc}
\end{figure}

The chronological results also showed substantial variation in achievable savings across releases. Release 11 was the most imbalanced test window, with only 258 quantifiable pairs among 2{,}343 promoted pairs. Its AUPRC fell to 0.382 against an 11.0\% positive rate, despite recall of 0.922 at a 50\% profiling budget (Fig.~\ref{fig:auprc}). Thus, the amount of profiling that can be withheld at a given retention target depends on the composition and prevalence of quantifiable responses in the incoming release rather than being a fixed property of the model.

The direct forest's out-of-fold scores on the promoted set had an expected calibration error of 0.013 and a Brier score of 0.099.

\subsection{The predictive signal generalizes across assay-mechanism families and an independent platform}

We next tested whether the screening-stage signal persisted when entire assay-mechanism families were excluded from model training. 
In leave-one-mechanism-family-out evaluation, a fingerprint-only model transferred poorly to held-out families, with a mean AUROC of 0.578. 
Adding the three-point screening response increased the mean AUROC to 0.862. 
Adding an explicit encoding of the assay-mechanism label on top of the screening features produced little further change, yielding a mean AUROC of 0.863.

Performance remained above chance for every held-out mechanism family, although the magnitude varied. 
AUROC was 0.966 for held-out kinase competition-binding assays, 0.863 for G-protein activation, 0.867 for co-factor recruitment, 0.813 for heterodimer assays, and 0.800 for Barr2 recruitment. 
Thus, screening-response information remained predictive even when the model had not been trained on examples from the test assay-mechanism family.

On the matched EPA ToxCast/Tox21 screens, 42{,}479 of 43{,}242 compound--endpoint pairs (98.2\%) were measured at a single screening concentration. 
Across 361 endpoints, ranking pairs by screening amplitude distinguished those that later yielded a potency from those that did not (AUROC $0.815 \pm 0.007$ with random cross-validation; $0.816 \pm 0.029$ with endpoints held out).
At a budget allowing multi-concentration profiling of 50\% of pairs, ranking by screening amplitude recovered approximately 91\% of pairs yielding a reportable potency estimate. 
Missed potencies were, again, concentrated among weak screening responses.

\section{Discussion}\label{sec12}

\subsection{Quantifiability complements activity-based hit selection}
This study distinguishes two outcomes of follow-up screening: observing activity and obtaining a reportable potency estimate. 
Among the 18{,}299 promoted pairs without a reported potency, 6{,}750 were nevertheless classified as active during 11-point profiling. 
Thus, activity confirmation alone would not have identified all experiments that failed to return the desired potency parameter. 
Predicting quantifiability addresses this additional decision problem: which already-promoted pairs are most likely to yield a reportable potency estimate under the current profiling workflow?

This objective complements, rather than replaces, activity prediction and false-positive filtering \parencite{Riniker2014,Boldini2024,Hesse2024}. 
Its contribution is the formulation and evaluation of an outcome relevant to experimental planning.

\subsection{Screening response and assay-dependent reportability}
The three-point response block was more informative than fingerprints alone and retained similar discrimination under scaffold-disjoint evaluation. 
Fingerprint-only performance deteriorated substantially under the same split, whereas combining response features with structural and contextual information remained useful. 
These comparisons support a substantial contribution from observed screening responses rather than molecular similarity alone.

Active-unquantifiable pairs had weaker response amplitudes than quantified pairs, and failure rates differed substantially across assay-mechanism families. 
This pattern is consistent with the dependence of potency estimation on response range, signal quality, and the concentrations tested \parencite{Sebaugh2011,Shockley2015,SrinivasanLloyd2024}. 
A low-amplitude response can still yield a relative potency if its transition and asymptotes are resolved \parencite{Sebaugh2011}.

Screening category and promotion reason are annotations generated during EvE Bio's primary-screening workflow. 
The full model could therefore partly reflect an association between these annotations and subsequent reporting outcomes. 
Removing both annotation groups left the AUROC essentially unchanged (difference $0.0005$; 95\% CI $-0.0001$ to $0.0011$). 
This result suggests that these annotations provided little additional discriminatory information beyond the remaining screening-stage features in the evaluated models.

\subsection{Practical value depends on the comparator and the decision rule}
In each of the eight release-ordered test windows, the direct forest recovered more quantifiable pairs than ranking by the reported screening maximum at a 50\% profiling budget, and it needed a smaller budget to retain 90\% of them. 

At 90\% retention the forest profiled 16{,}780 of the 30{,}120 pairs, against 18{,}724 for ranking by the reported screening maximum. 
That is 1{,}944 fewer experiments, a 10.4\% reduction relative to the activity-ranking baseline, and 13{,}340 fewer than profiling every promoted pair. 
An unquantifiable profile is not necessarily uninformative: it can still indicate weak activity, bear on selectivity, or show that a different assay format or concentration range is needed. 
Quantifiability should therefore be weighed together with chemical diversity, target relevance, and what is lost by leaving a given compound--assay pair unprofiled.

\subsection{Generalization and limits of retrospective evaluation}
Scaffold-disjoint and mechanism-held-out analyses test transfer within EvE Bio.
In the release-ordered evaluation, each model was fit, and each historical summary computed, using earlier releases only \parencite{Sheridan2013}. 
Every screening-stage field, including the reported screening maximum, screening category, and promotion reason, was recorded before 11-point profiling (Supplementary Table S1). 
No information from the test release, including the test pair's profiling outcome, entered the historical summaries.

\subsection{Limitations}
All EvE Bio results are conditional on historical promotion: only pairs that were actually profiled have outcomes, so performance among pairs that were never profiled is unknown. 
Quantifiability indicates whether a potency estimate was reported under the assay and analysis procedures used here. 
It does not measure the accuracy of that estimate, its therapeutic value, or any fixed property of the molecule.

Several features of the evaluation limit how far the results extend. 
Random-split results use historical summaries computed from the full screening matrix, so they describe interpolation among pairs already in the dataset, not performance on a future release. 
In ToxCast/Tox21, we examined only the association between screening amplitude and a reportable potency estimate from subsequent profiling; we did not test whether the trained EvE Bio model or its profiling-budget estimates transfer. 
Finally, impurity importance and the nested feature gains are predictive summaries, not causal effects, and the estimated reductions in experiment counts do not establish realized monetary savings.

\FloatBarrier

\section*{Funding}
The author did not receive support from any organization for the submitted work.

\section*{Declaration of competing interest}
The author declares no competing interests.

\section*{Ethics approval}
Not applicable. This study analyzed publicly available screening data and did not involve human participants or animals.

\section*{Consent to participate}
Not applicable.

\section*{Consent for publication}
Not applicable.

\section*{Data availability}
The data analyzed in this study were obtained from two publicly available sources. The primary analyses used EvE Bio Data Releases \#1--\#11, comprising compound--assay three-point screening and 11-point concentration--response data, from a local snapshot of the release-11 bundle dated 15 May 2026. These data are publicly available through the EvE Bio Data portal \parencite{EvEBioPortal2026} under the Creative Commons CC BY-NC-SA 4.0 license. The independent assessment of the association between screening amplitude and quantifiability used publicly available U.S. Environmental Protection Agency ToxCast/Tox21 high-throughput screening and concentration--response data from the invitroDB database (version 4.3) \parencite{EPAInvitrodb2025}. No new experimental data were generated in this study.

\section*{Materials availability}
Not applicable.

\section*{Code availability}
Analysis code, including feature construction and the evaluation scripts, is available at \url{https://github.com/sean6211/drug-response-quantifiability}.

\section*{CRediT authorship contribution statement}
\textbf{Sean Lim:} Conceptualization, Methodology, Formal analysis, Visualization, Writing -- original draft, Writing -- review \& editing.

\section*{Preprint disclosure}
A preprint of this manuscript is available at \texttt{arXiv:2608.26538} (\url{https://arxiv.org/abs/2608.26538}).

\section*{Supplementary material}
Supplementary Tables S1 and S2 are supplied in the separate file \path{JPTM_Supplementary_Tables.pdf}.

\section*{Acknowledgements}
The author thanks EvE Bio, LLC for making its compound--assay screening and concentration--response data publicly available, and the U.S. Environmental Protection Agency and the Tox21 consortium for making the ToxCast/Tox21 high-throughput screening data available through the invitroDB database.

\printbibliography[title={References}]

@article{Hughes2011,
  author  = {Hughes, J. P. and Rees, S. and Kalindjian, S. B. and Philpott, K. L.},
  title   = {Principles of Early Drug Discovery},
  journal = {British Journal of Pharmacology},
  year    = {2011},
  volume  = {162},
  number  = {6},
  pages   = {1239--1249},
  doi     = {10.1111/j.1476-5381.2010.01127.x}
}

@article{SrinivasanLloyd2024,
  author  = {Srinivasan, Bharath and Lloyd, Matthew D.},
  title   = {Dose--Response Curves and the Determination of {IC50} and {EC50} Values},
  journal = {Journal of Medicinal Chemistry},
  year    = {2024},
  volume  = {67},
  number  = {20},
  pages   = {17931--17934},
  doi     = {10.1021/acs.jmedchem.4c02052}
}

@article{Buterez2023,
  author  = {Buterez, David and Janet, Jon Paul and Kiddle, Steven J. and Li{\`o}, Pietro},
  title   = {{MF-PCBA}: Multifidelity High-Throughput Screening Benchmarks for Drug Discovery and Machine Learning},
  journal = {Journal of Chemical Information and Modeling},
  year    = {2023},
  volume  = {63},
  number  = {9},
  pages   = {2667--2678},
  doi     = {10.1021/acs.jcim.2c01569}
}

@article{Riniker2014,
  author  = {Riniker, Sereina and Wang, Yuan and Jenkins, Jeremy L. and Landrum, Gregory A.},
  title   = {Using Information from Historical High-Throughput Screens to Predict Active Compounds},
  journal = {Journal of Chemical Information and Modeling},
  year    = {2014},
  volume  = {54},
  number  = {7},
  pages   = {1880--1891},
  doi     = {10.1021/ci500190p}
}

@article{Baell2010,
  author  = {Baell, Jonathan B. and Holloway, Georgina A.},
  title   = {New Substructure Filters for Removal of Pan Assay Interference Compounds ({PAINS}) from Screening Libraries and for Their Exclusion in Bioassays},
  journal = {Journal of Medicinal Chemistry},
  year    = {2010},
  volume  = {53},
  number  = {7},
  pages   = {2719--2740},
  doi     = {10.1021/jm901137j}
}

@article{Dahlin2015,
  author  = {Dahlin, Jayme L. and Nissink, J. Willem M. and Strasser, Jessica M. and Francis, Subhashree and Higgins, LeeAnn and Zhou, Hui and Zhang, Zhiguo and Walters, Michael A.},
  title   = {{PAINS} in the Assay: Chemical Mechanisms of Assay Interference and Promiscuous Enzymatic Inhibition Observed during a Sulfhydryl-Scavenging {HTS}},
  journal = {Journal of Medicinal Chemistry},
  year    = {2015},
  volume  = {58},
  number  = {5},
  pages   = {2091--2113},
  doi     = {10.1021/jm5019093}
}

@article{Tan2024,
  author  = {Tan, Lu and Hirte, Steffen and Palmacci, Vincenzo and Stork, Conrad and Kirchmair, Johannes},
  title   = {Tackling Assay Interference Associated with Small Molecules},
  journal = {Nature Reviews Chemistry},
  year    = {2024},
  volume  = {8},
  number  = {5},
  pages   = {319--339},
  doi     = {10.1038/s41570-024-00593-3}
}

@article{Malo2010,
  author  = {Malo, Nathalie and Hanley, James A. and Carlile, Graeme and Liu, Jing and Pelletier, Jerry and Thomas, David and Nadon, Robert},
  title   = {Experimental Design and Statistical Methods for Improved Hit Detection in High-Throughput Screening},
  journal = {Journal of Biomolecular Screening},
  year    = {2010},
  volume  = {15},
  number  = {8},
  pages   = {990--1000},
  doi     = {10.1177/1087057110377497}
}

@article{Sebaugh2011,
  author  = {Sebaugh, J. L.},
  title   = {Guidelines for Accurate {EC50}/{IC50} Estimation},
  journal = {Pharmaceutical Statistics},
  year    = {2011},
  volume  = {10},
  number  = {2},
  pages   = {128--134},
  doi     = {10.1002/pst.426}
}

@article{Shockley2015,
  author  = {Shockley, Keith R.},
  title   = {Quantitative High-Throughput Screening Data Analysis: Challenges and Recent Advances},
  journal = {Drug Discovery Today},
  year    = {2015},
  volume  = {20},
  number  = {3},
  pages   = {296--300},
  doi     = {10.1016/j.drudis.2014.10.005}
}

@article{Sheridan2013,
  author  = {Sheridan, Robert P.},
  title   = {Time-Split Cross-Validation as a Method for Estimating the Goodness of Prospective Prediction},
  journal = {Journal of Chemical Information and Modeling},
  year    = {2013},
  volume  = {53},
  number  = {4},
  pages   = {783--790},
  doi     = {10.1021/ci400084k}
}

@article{Feshuk2023,
  author  = {Feshuk, M. and Kolaczkowski, L. and Dunham, K. and Davidson-Fritz, S. E. and Carstens, K. E. and Brown, J. and Judson, R. S. and Paul Friedman, K.},
  title   = {The {ToxCast} Pipeline: Updates to Curve-Fitting Approaches and Database Structure},
  journal = {Frontiers in Toxicology},
  year    = {2023},
  volume  = {5},
  eid     = {1275980},
  doi     = {10.3389/ftox.2023.1275980}
}

@article{Macarron2011,
  author  = {Macarron, Ricardo and Banks, Martyn N. and Bojanic, Dejan and Burns, David J. and Cirovic, Dragan A. and Garyantes, Tina and Green, Darren V. S. and Hertzberg, Robert P. and Janzen, William P. and Paslay, Jeff W. and Schopfer, Ulrich and Sittampalam, G. Sitta},
  title   = {Impact of High-Throughput Screening in Biomedical Research},
  journal = {Nature Reviews Drug Discovery},
  year    = {2011},
  volume  = {10},
  number  = {3},
  pages   = {188--195},
  doi     = {10.1038/nrd3368}
}

@article{Inglese2006,
  author  = {Inglese, James and Auld, Douglas S. and Jadhav, Ajit and Johnson, Ronald L. and Simeonov, Anton and Yasgar, Adam and Zheng, Wei and Austin, Christopher P.},
  title   = {Quantitative High-Throughput Screening: A Titration-Based Approach that Efficiently Identifies Biological Activities in Large Chemical Libraries},
  journal = {Proceedings of the National Academy of Sciences of the United States of America},
  year    = {2006},
  volume  = {103},
  number  = {31},
  pages   = {11473--11478},
  doi     = {10.1073/pnas.0604348103}
}

@article{Zhang1999,
  author  = {Zhang, Ji-Hu and Chung, Thomas D. Y. and Oldenburg, Kevin R.},
  title   = {A Simple Statistical Parameter for Use in Evaluation and Validation of High Throughput Screening Assays},
  journal = {Journal of Biomolecular Screening},
  year    = {1999},
  volume  = {4},
  number  = {2},
  pages   = {67--73},
  doi     = {10.1177/108705719900400206}
}

@article{Vamathevan2019,
  author  = {Vamathevan, Jessica and Clark, Dominic and Czodrowski, Paul and Dunham, Ian and Ferran, Edgardo and Lee, George and Li, Bin and Madabhushi, Anant and Shah, Parantu and Spitzer, Michaela and Zhao, Shanrong},
  title   = {Applications of Machine Learning in Drug Discovery and Development},
  journal = {Nature Reviews Drug Discovery},
  year    = {2019},
  volume  = {18},
  number  = {6},
  pages   = {463--477},
  doi     = {10.1038/s41573-019-0024-5}
}

@article{Mayr2018,
  author  = {Mayr, Andreas and Klambauer, G{\"u}nter and Unterthiner, Thomas and Steijaert, Marvin and Wegner, J{\"o}rg K. and Ceulemans, Hugo and Clevert, Djork-Arn{\'e} and Hochreiter, Sepp},
  title   = {Large-Scale Comparison of Machine Learning Methods for Drug Target Prediction on {ChEMBL}},
  journal = {Chemical Science},
  year    = {2018},
  volume  = {9},
  number  = {24},
  pages   = {5441--5451},
  doi     = {10.1039/c8sc00148k}
}

@article{RogersHahn2010,
  author  = {Rogers, David and Hahn, Mathew},
  title   = {Extended-Connectivity Fingerprints},
  journal = {Journal of Chemical Information and Modeling},
  year    = {2010},
  volume  = {50},
  number  = {5},
  pages   = {742--754},
  doi     = {10.1021/ci100050t}
}

@article{BemisMurcko1996,
  author  = {Bemis, Guy W. and Murcko, Mark A.},
  title   = {The Properties of Known Drugs. 1. {Molecular frameworks}},
  journal = {Journal of Medicinal Chemistry},
  year    = {1996},
  volume  = {39},
  number  = {15},
  pages   = {2887--2893},
  doi     = {10.1021/jm9602928}
}

@article{WallachHeifets2018,
  author  = {Wallach, Izhar and Heifets, Abraham},
  title   = {Most Ligand-Based Classification Benchmarks Reward Memorization Rather than Generalization},
  journal = {Journal of Chemical Information and Modeling},
  year    = {2018},
  volume  = {58},
  number  = {5},
  pages   = {916--932},
  doi     = {10.1021/acs.jcim.7b00403}
}

@article{TruchonBayly2007,
  author  = {Truchon, Jean-Fran{\c{c}}ois and Bayly, Christopher I.},
  title   = {Evaluating Virtual Screening Methods: Good and Bad Metrics for the ``Early Recognition'' Problem},
  journal = {Journal of Chemical Information and Modeling},
  year    = {2007},
  volume  = {47},
  number  = {2},
  pages   = {488--508},
  doi     = {10.1021/ci600426e}
}

@article{Pedregosa2011,
  author  = {Pedregosa, Fabian and Varoquaux, Ga{\"e}l and Gramfort, Alexandre and Michel, Vincent and Thirion, Bertrand and Grisel, Olivier and Blondel, Mathieu and Prettenhofer, Peter and Weiss, Ron and Dubourg, Vincent and Vanderplas, Jake and Passos, Alexandre and Cournapeau, David and Brucher, Matthieu and Perrot, Matthieu and Duchesnay, {\'E}douard},
  title   = {{Scikit-learn}: Machine Learning in {Python}},
  journal = {Journal of Machine Learning Research},
  year    = {2011},
  volume  = {12},
  pages   = {2825--2830}
}

@software{RDKit2026,
  title        = {{RDKit}: Open-source cheminformatics},
  shorttitle   = {{RDKit}},
  year         = {2026},
  version      = {2026.03.2},
  entrysubtype = {Computer software},
  url          = {https://www.rdkit.org},
  urldate      = {2026-10-29}
}

@article{Breiman2001,
  author  = {Breiman, Leo},
  title   = {Random forests},
  journal = {Machine Learning},
  year    = {2001},
  volume  = {45},
  number  = {1},
  pages   = {5--32},
  doi     = {10.1023/A:1010933404324}
}

@article{Boldini2024,
  author  = {Boldini, Davide and Friedrich, Lukas and Kuhn, Daniel and Sieber, Stephan A.},
  title   = {Machine Learning Assisted Hit Prioritization for High Throughput Screening in Drug Discovery},
  journal = {ACS Central Science},
  year    = {2024},
  volume  = {10},
  number  = {4},
  pages   = {823--832},
  doi     = {10.1021/acscentsci.3c01517}
}

@article{Hesse2024,
  author  = {Hesse, Joshua and Boldini, Davide and Sieber, Stephan A.},
  title   = {Machine Learning-Driven Data Valuation for Optimizing High-Throughput Screening Pipelines},
  journal = {Journal of Chemical Information and Modeling},
  year    = {2024},
  volume  = {64},
  number  = {21},
  pages   = {8142--8152},
  doi     = {10.1021/acs.jcim.4c01547}
}

@dataset{EvEBioPortal2026,
  author       = {{EvE Bio, LLC}},
  title        = {{EvE Bio} data releases \#1--\#14},
  year         = {2026},
  url          = {https://data.evebio.org/},
  entrysubtype = {Data set},
  note         = {Local release-11 snapshot dated 15 May 2026. Only releases \#1--\#11 were analyzed in this study}
}

@dataset{EPAInvitrodb2025,
  author       = {{U.S. Environmental Protection Agency}},
  title        = {{ToxCast invitrodb} Version 4.3},
  year         = {2025},
  url          = {https://doi.org/10.23645/epacomptox.6062623.v14},
  entrysubtype = {Data set},
  note         = {August 2025 database release},
  doi          = {10.23645/epacomptox.6062623.v14}
}

\end{document}


{\Large\bfseries Supplementary tables\par}
\medskip
{\large Predicting Dose--Response Success from Primary Screening Data\par}
\medskip
Sean Lim\\
Department of BioSciences, Rice University, Houston, TX, USA\\
Correspondence: \href{mailto:sean78623@gmail.com}{sean78623@gmail.com}
\medskip

\section*{Supplementary Table S1}
\label{sec:s1}

\noindent
Timing of fields in the local release-11 snapshot dated 15 May 2026. Screening-stage fields are populated for pairs that were not profiled. The reported screening maximum agrees with the three-point screen and not with the 11-point maximum.

\begin{center}
\small
\setlength{\tabcolsep}{4pt}
\renewcommand{\arraystretch}{1.15}
\begin{tabular}{@{}>{\raggedright\arraybackslash}p{0.24\textwidth}>{\raggedright\arraybackslash}p{0.68\textwidth}@{}}
\toprule
Field & What the snapshot shows \\
\midrule
Three-point features &
Computed from observed points in the Screening (3pt) phase only. \\
Screening maximum &
Pearson $r = 0.999985$ with the maximum of the three-point concentration means (median absolute difference 0.25; 95.8\% equal that maximum after rounding). Pearson $r = 0.164$ with the 11-point maximum. \\
Screening category &
Non-null for 433{,}627 pairs that were not promoted and have no profiling maximum. Equal to the final category for 0\% of the 32{,}971 promoted pairs. \\
Promotion reason &
Recorded for every promoted pair as the reason the pair was sent to profiling. \\
Historical summaries &
Calculated from the full screening matrix in the random-split analyses, and recomputed from earlier releases in the release-ordered evaluation. \\
Profiling outcomes &
The 11-point maximum, pXC$_{50}$, and final category come from the profiling experiment. The profiling maximum is null for all 549{,}578 pairs that were not promoted. These fields were outcomes, not predictors. \\
\bottomrule
\end{tabular}
\end{center}

\clearpage
\section*{Supplementary Table S2}
\label{sec:s2}

\noindent
RDKit descriptors for quantifiable and active-unquantifiable groups. Values are mean $\pm$ SD. Cohen's $d$ is quantifiable minus active-unquantifiable over the pooled SD. Pair rows use the 14{,}412 quantifiable and 6{,}602 active-unquantifiable pairs with a parseable SMILES, and the 95\% interval for $d$ is from 2{,}000 compound-cluster bootstrap resamples. Compound rows count each compound once per group (726 and 671 compounds; 566 appear in both). Welch degrees of freedom were 10{,}357--11{,}531 for pairs and 1{,}271--1{,}380 for compounds. MW, molecular weight; HBD and HBA, hydrogen-bond donors and acceptors; TPSA, topological polar surface area.

\begin{center}
\small
\setlength{\tabcolsep}{3.5pt}
\renewcommand{\arraystretch}{1.15}
\begin{adjustbox}{max width=\textwidth}
\begin{tabular}{@{}lcccccc@{}}
\toprule
Descriptor & Quantifiable & Active, unquantifiable & Welch $t$ & $p$ & $d$ & 95\% CI for $d$ \\
\midrule
\multicolumn{7}{@{}l}{\textit{Pairs}} \\
MW & $524.7 \pm 239.5$ & $544.7 \pm 270.1$ & $-5.16$ & $2.6 \times 10^{-7}$ & $-0.080$ & $-0.154$, $-0.004$ \\
LogP & $3.35 \pm 2.46$ & $3.41 \pm 2.96$ & $-1.61$ & $0.11$ & $-0.026$ & $-0.127$, $0.077$ \\
HBD & $2.75 \pm 2.69$ & $3.10 \pm 3.48$ & $-7.27$ & $3.7 \times 10^{-13}$ & $-0.119$ & $-0.189$, $-0.039$ \\
HBA & $7.12 \pm 4.46$ & $7.39 \pm 5.04$ & $-3.71$ & $2.1 \times 10^{-4}$ & $-0.058$ & $-0.139$, $0.029$ \\
Rotatable bonds & $6.45 \pm 4.33$ & $7.19 \pm 4.88$ & $-10.59$ & $4.2 \times 10^{-26}$ & $-0.165$ & $-0.251$, $-0.077$ \\
TPSA & $116.6 \pm 82.1$ & $125.7 \pm 100.5$ & $-6.47$ & $1.0 \times 10^{-10}$ & $-0.104$ & $-0.173$, $-0.026$ \\
\midrule
\multicolumn{7}{@{}l}{\textit{Compounds}} \\
MW & $466.2 \pm 233.1$ & $475.7 \pm 244.9$ & $-0.74$ & $0.46$ & $-0.040$ & --- \\
LogP & $3.13 \pm 2.79$ & $3.04 \pm 3.55$ & $0.54$ & $0.59$ & $0.029$ & --- \\
HBD & $2.46 \pm 3.04$ & $2.58 \pm 3.19$ & $-0.73$ & $0.46$ & $-0.039$ & --- \\
HBA & $5.82 \pm 4.14$ & $6.15 \pm 4.87$ & $-1.37$ & $0.17$ & $-0.074$ & --- \\
Rotatable bonds & $6.56 \pm 5.32$ & $6.64 \pm 5.45$ & $-0.26$ & $0.79$ & $-0.014$ & --- \\
TPSA & $101.0 \pm 89.8$ & $107.3 \pm 98.9$ & $-1.24$ & $0.22$ & $-0.067$ & --- \\
\bottomrule
\end{tabular}
\end{adjustbox}
\end{center}